\documentclass[12pt,a4paper]{cibb}

\makeatletter
\providecommand{\@ordinalM}[2]{#1}
\makeatother

\usepackage{subfigure,graphicx}
\usepackage{amsmath,amsfonts,latexsym,amssymb,euscript,xr}
\usepackage{booktabs}
\usepackage[nodayofweek]{datetime}
\usepackage{hyperref}
\usepackage{fmtcount}
\usepackage[english]{datenumber}
\usepackage[absolute]{textpos}
\usepackage{ulem}
\usepackage{comment}

\usepackage{makecell}
\usepackage{multirow}
\usepackage[table,xcdraw]{xcolor}
\usepackage{tabularx}
\usepackage{truncate}

\usepackage[table]{xcolor}
\usepackage{color,colortbl,tabularx}

\usepackage{algorithm}
\usepackage{algpseudocode}
\usepackage{amsmath}

\usepackage[english]{babel}
\usepackage[protrusion=true,expansion=true]{microtype}
\usepackage{amsmath,amsfonts,amsthm}
\usepackage{pifont}

\definecolor{LightBlue}{rgb}{0.88,0.9,0.9}

\title{\Large \textbf{Expert Knowledge \& Machine Understanding: Bridging Reactome’s Ontology with LLM Semantic Embeddings}
}

\author{\large Susanna Bravi$^{1+}$, Riccardo De Luca$^{1,2+}$, Rosa Sicilia$^3$, Christine Nardini$^1$, Mario Santoro$^{1*}$}

\address{\footnotesize $\ $\\$^1$ Istituto per le Applicazioni del Calcolo Mauro Picone, Italian National Research Council, Rome, Italy.\\
$^2$ Unit of Artificial Intelligence and Computer Systems, Department of Engineering, Università Campus Bio-Medico di Roma, Rome, Italy \\
$^3$ UniCamillus-Saint Camillus International University of Health Sciences, Rome, Italy \\
\bigskip
ORCID codes: SB 0009-0003-8269-5861; RDL 0009-0008-5058-6309; RS 0000-0002-2513-0827; CN 0000-0001-7601-321X; MS 0000-0001-6626-9430.
\bigskip
\newline
$^*$corresponding author: mario-santoro@cnr.it
\newline
$^+$ equally contributing
}

\abstract{\small knowledgebase, Reactome, LLM, hierarchy, embedding. \normalsize
\\[17pt]
{\bf Abstract 
}
Biological knowledgebases like Reactome provide high-quality pathways that include biological elements' relationships and textual descriptions (metadata). The quality of such pathways is granted by manual curation, that presents, however, significant scalability challenges. 
Lately, numerous NLP tools have been proposed to cope with this issue, leveraging textual information to automatically expand biological knowledgebases. However, little exploration has been done so far to assess whether relationships among textual descriptions mirror higher order biological relationships. 
This study explores whether human-written descriptions in Reactome can be used to infer the experts' defined global hierarchical structure.
To test this, we extracted from Reactome the Homo Sapiens hierarchy of pathways and their reactions (\textit{Reactome Hierarchy}), and used textual metadata to reconstruct a \textit{Semantic Hierarchy}, combining a sentence transformer model (SPECTER2) with a modified agglomerative nesting algorithm and a graph reconstruction algorithm.
Quantitative (Laplacian Spectral Distance and Bootstrapping) and qualitative (global topological metrics) analyses confirm our hypothesis and indicate that the global hierarchical structure of pathways can be inferred by experts textual metadata.}

\begin{document}

\renewcommand{\thefootnote}{}
\footnotetext{\small{Article version: \datedate $\;$ h\currenttime  $\;$ CET}}

\thispagestyle{myheadings}
\pagestyle{myheadings}
\markright{\tt Proceedings of CIBB 2026}

\section{Introduction}
\label{sec:SCIENTIFIC-BACKGROUND}
Biological pathways are essential conceptual tools for mapping the complex biochemical interactions that drive biological processes. To manage this vast amount of information (molecules and associated relations) they are often collected in comprehensive knowledgebases like Reactome \cite{ragueneau2026reactome} that ensures high quality and structured pathway data by manual curation from the primary literature. In Reactome, a Pathway is defined as a series of interconnected Reactions, based on biological relevance. 
Reactome \textit{Reactions} are the basic units of this knowledgebase, defined as the ensemble of everyday language \textit{reactions} (binding, phosphorylation, molecular transport, etc.) and the involved molecules. Throughout this paper the capital letter distinguishes the two concepts.
The Reactions are then aggregated to form Pathways, which are further organized hierarchically into a large interconnected graph that can be accessed using the Neo4j Reactome Graph Database \cite{fabregat2018reactomegraph}, here named \textit{Reactome Hierarchy (RH)}. In addition to strictly structured information, Reactome contains rich textual metadata. In fact, each Pathway and the Reactions that compose it, are provided with a title and a descriptive human-written summary of their purpose.

Although manual curation guarantees high fidelity, it presents a significant scalability challenge. Recent advances in Natural Language Processing (NLP) and Large Language Models (LLMs) suggest that textual metadata may offer a route toward more automated knowledge integration and interaction.

Maeda et al. \cite{maeda2023automatic} employed GPT-derived models to translate natural language descriptions of the metadata into Antimony strings, a domain-specific markup language for biological models, which were then converted into executable kinetic models. Wu et al. \cite{wu2025application} instead, advanced computer-assisted curation by proposing a pipeline that maps new genes to existing pathways and refines functional annotations through literature-supported entity extraction. Regarding the interaction, Mohammadi et al. \cite{mohammadi2025react} developed \textit{React-to-Me}, a conversational assistant that combines hybrid retrieval-augmented generation (RAG) with constrained decoding to ground user queries in Reactome content. These tools demonstrate that textual metadata can support the retrieval of structured biological information.

In this study, we investigate the Reactome case to assess how well semantic embeddings derived solely from human-written annotations' titles and descriptions can reconstruct Reactome's biological hierarchy. 
Complementing prior efforts focused on kinetic modeling, conversational interfaces, or new literature ingestion, our contribution is twofold: (1) we focus exclusively on \textit{structural reconstruction}, measuring how semantic similarity among existing textual metadata alone can recover hierarchical relationships, (2) we introduce a graph-theoretic evaluation framework to quantify the topological alignment between the resulting \textit{Semantic Hierarchy} (SH) and the expert-curated RH. By isolating the informational signal within pre-existing curated annotations, this work provides a preliminary and scalable opening towards text-driven pathway reconstruction.


\section{Data and Methods}
\label{sec:DATA-AND-METHODS}

The proposed approach is structured in four phases: (i) we extract the RH from the Reactome database Version V95 December 2025 \cite{ragueneau2026reactome}(Section \ref{subsec:reactome-hierarchy}); (ii) we build the semantic embeddings of Pathways and Reactions textual descriptions (Section \ref{sec:semantic-embeddings}); (iii) we use the generated embeddings to build a SH, by combining a clustering technique (Section \ref{sec:HIERARCHICAL-CLUSTERING}) with a graph reconstruction algorithm (Section \ref{sec:cluster2graph}), (iv) we compare SH to the reference RH (Section \ref{sec:hiercomp}).

\subsection{Reactome Hierarchy Retrieval}
\label{subsec:reactome-hierarchy}
To extract the RH we queried the database filtering for \textit{Homo sapiens} using it as the reference species given its popularity and renown quality in the scientific reference community.
We then retrieved all Pathway and Reaction nodes, along with their hierarchical relationships and textual descriptions. 
In Reactome, Pathways and Reactions are categorized as \textit{Event} nodes and further divided into subclasses to represent specific biological scenarios; we included all such subclasses in our retrieval. We aggregated all elements into a graph with textual description as node attributes to support downstream semantic embedding (see Section \ref{sec:semantic-embeddings}).

\subsection{Semantic Embedding of Description Database and Data Pre-processing}
\label{sec:semantic-embeddings}

Based on RH, we created a tabular dataset of textual descriptions extracting for each node: Reactome ID,  entity type (class or subclass), title and textual description. A sample of this dataset can be seen in Table \ref{tab:sentence-dataset-sample}.

\begin{table}[httb!] \footnotesize
\centering
\newcommand{\mytabtruncate}[1]{\truncate{\hsize}{#1}}
\begin{tabularx}{0.9\textwidth}{ l  l  l  X }

\toprule
\textbf{Reactome ID} & \textbf{Type} & \textbf{Name} & \textbf{Description} \\
\midrule

\rowcolor{LightBlue}
R-HSA-9612973 & TopLevelPathway & Autophagy & \mytabtruncate{Autophagy is an intracellular degradation process that is triggered by cellular stresses. There are three primary types of autophagy - macroautophagy, chaperone-mediated autophagy (CMA) and late endosomal microautophagy.} \\

R-HSA-9615710 & Pathway & Late endosomal & \mytabtruncate{Microautophagy (MI) is a non-selective autophagic pathway that involves internalisation of cytosolic cargo through invaginations of the lysosomal membrane.} \\

\rowcolor{LightBlue}
R-HSA-1632852 & Pathway & Macroautophagy & \mytabtruncate{Macroautophagy (hereafter referred to as autophagy) acts as a buffer against starvation by liberating building materials and energy sources from cellular components.} \\

R-HSA-9613829 & Pathway & \makecell[l]{Chaperone \\ Mediated Autophagy} & \mytabtruncate{In contrary to the vesicle-mediated macroautophagy, the chaperone mediated mechanism of autophagy selectively targets individual proteins to the lysosome for degradation.} \\

\rowcolor{LightBlue}
R-HSA-9626253 & Reaction & \makecell[l]{HSPA8 binds \\ LAMP2a multimers} & \mytabtruncate{Intracellular proteins are targeted for proteolytic degradation in lysosome with the aid of chaperones. Heat shock cognate 71 kDa protein (HSPA8) transports substrates from the cytosol.} \\

R-HSA-5682388 & BlackBoxEvent & \makecell[l]{Autophagosome \\ maturation} & \mytabtruncate{The mechanisms involved in the closure of the phagophore into an enclosed autophagosome are poorly understood.} \\
\bottomrule

\end{tabularx}
\caption{\textbf{Sample of the description dataset}. Dataset used to generate the sentence embeddings, concatenating Name and Description column.}
\label{tab:sentence-dataset-sample}
\end{table}

For SH, we employed SPECTER2 \cite{singh2023scirepevalmultiformatbenchmarkscientific}, a sentence transformer model pre-trained on titles and abstracts of scientific publications using contrastive learning on a scientific citation graph. Consequently, it is better suited for the biomedical domain than general-purpose model.

Pre-processing includes a rigorous cleaning process and removal of bibliographic citations and the concatenation of node name and description with $[SEP]$ to reduce number of duplicate descriptions. 
For entities with descriptions exceeding the maximum token limit ($512$), we splitted it in sentences using  the $12l$-layer (\textit{sat-12l-sm}) model in the SaT (Segment any Text) library \cite{frohmann2024segmenttextuniversalapproach}
to preserve sentence boundaries while incorporating $10\%$ overlap to guarantee contextual continuity.

\subsection{Modified Hierarchical Clustering } 
\label{sec:HIERARCHICAL-CLUSTERING}

To construct the SH, we extend an Agglomerative Nesting (AGNES) clustering algorithm\cite{Wiley}, to allow for the simultaneous merging of multiple nodes which cannot be performed by standard binary hierarchical clustering.
This approach is named in the following multiway hierarchy. 

A standard linkage matrix stores all pairwise distances between the initial data points (Pathways and Reaction-Like Events). Then, to support a non-binary hierarchy, for each node the difference between the current merge height and the height at which the node was originally formed is evaluated. 
For differences smaller than a data-driven threshold $\epsilon$, ( minimum distance different from zero in the linkage matrix), the node is absorbed and its children are directly attached to the new parent; otherwise, the node is retained as an intermediate branch. A new parent node is then created to group the resulting children.

The linkage distances between the newly formed cluster and all remaining active clusters are updated iteratively using the Lance-Williams formula. 

Although the algorithm generates a multi-way hierarchical tree, the resulting topology is not yet suitable for direct comparison with the RH, since the newly generated clusters are mathematical aggregation nodes (named \textit{Clusters} in the following), but may not represent true pathways.
To bridge this structural gap, we implemented the following reconstruction strategy.

\subsection{Cluster to Graph Reconstruction}
\label{sec:cluster2graph}



To replace Clusters with the most appropriate Pathway nodes, thereby connecting the corresponding Reactions and Sub-Pathways directly to their correct parents, we explore the tree using a post-order depth-first search (DFS), visiting descendant nodes before their parent nodes. This bottom-up approach ensures that, at each iteration, the Cluster node under evaluation is connected exclusively to Pathway or Reaction nodes, as nested sub-clusters have already been resolved and removed in earlier iterations. For each Cluster, three distinct scenarios can occur: \textbf{(i) Only Reactions are present}, in which case they are linked directly to the parent Cluster; \textbf{(ii) At most one Pathway is present}, in which case any Reactions become its children, and it is linked directly to the parent Cluster; \textbf{(iii) Multiple Pathways coexist}, in which case hierarchical relationships among Pathways and Reactions are resolved using a Reactome-derived ancestor--descendant mapping, where each node is assigned to its closest valid ancestor among its siblings, while nodes with no such ancestor are linked directly to the parent Cluster.

At this stage, all children of the Cluster node have been sorted and linked to the Cluster parent. The current Cluster node is then deleted and the algorithm moves to the next one according to the previously defined order.

\subsection{Hierarchy Comparison}
\label{sec:hiercomp}
To quantify the similarity between SH and RH we used Edit Distance, based on the Frobenius norm between adjacency matrices, to quantify edge-level changes; and Laplacian Spectral Distance \cite{banerjee2009structuraldistanceevolutionaryrelationship} (LSD in $[0,\ln(2)]$), computed from the normalized graph Laplacian, to assess global topological divergence.
We first computed global topological measures to qualitatively explore their order of magnitude in the two hierarchies.  
Then, we assessed the significance of the observed LSD by bootstrapping. We generated 500 null models by globally rewiring SH while preserving the degree distribution, and computed the LSD between each rewired graph and RH. 

\section{Results}
\label{sec:RESULTS}
RH is composed of $18,732$ nodes ($2,848$ Pathways and $15,884$ Reactions) and $19,613$ edges. Node calculation includes all subclasses of Pathway and Reaction. The nodes distribution by subclass for the retrieved RH is summarized in Table \ref{tab:nodes-per-type}. The 29 nodes classified as \textit{TopLevelPathway} denote high-level biological themes (e.g., Cell Cycle, Immune System, and Metabolism), and serve as roots of the RH and macro-level anchors for hierarchical alignment evaluation.

To construct SH, high-dimensional embeddings were generated for each entity, resulting in a feature matrix of size $[18,732 \times 768]$ that served as the primary input for the clustering and hierarchical construction phases. The result is a graph with the same number of nodes as RH ($18,732$), but fewer edges ($18,703$). 
Computations were performed on a  node equipped with 64 Intel Xeon Gold 6883 CPU cores (2.00 GHz) and 4 NVIDIA A30 GPUs with 24 GB of memory each. This corresponds to the following computational times: embedding generation processed 18,732 descriptions in $\sim 2m$ ($\sim 167 row/s$), and clustering completed in $\sim 46s$ ($\sim 407 row/s$), graph reconstruction required $\sim 6s$ ($\sim 6,230 node/s$).


\begin{table}[httb!] \small
\centering
\begin{tabularx}{0.9\textwidth}{X r @{\hskip 3em} X r}
\toprule
\textbf{Class / Subclass} & \textbf{Nodes} & \textbf{Class / Subclass} & \textbf{Nodes} \\
\midrule
\rowcolor{LightBlue}
\textit{Pathway} (superclass) & 2{,}807 & \textit{ReactionLikeEvent} (superclass) & 0 \\
\quad CellLineagePath & 12 & \quad BlackBoxEvent & 2{,}894 \\
\rowcolor{LightBlue}
\quad TopLevelPathway & 29 & \quad CellDevelopmentStep & 22 \\
& & \quad Depolymerisation & 6 \\
\rowcolor{LightBlue}
& & \quad FailedReaction & 454 \\
& & \quad Polymerisation & 43 \\
\rowcolor{LightBlue}
& & \quad Reaction & 12{,}465 \\
\midrule
\textbf{Total} & \textbf{2{,}848} & \textbf{Total} & \textbf{15{,}884} \\
\bottomrule
\end{tabularx}
\caption{\textbf{Node composition of the Reactome Hyerarchy (RH) by ontological class.} Distribution of nodes across the two primary ontological classes, \textit{Pathway} and \textit{ReactionLikeEvent}, and their respective subclasses in the Reactome Graph Database, filtered to retain only \textit{Homo sapiens}-related nodes. }
\label{tab:nodes-per-type}
\end{table}

The structural alignment between the reconstructed SH and the reference RH was validated through both local and global graph comparison metrics, yielding highly robust results. The local structural divergence, measured by the Frobenius distance, was $216.88$; even though this reflects specific edge-level variations between the SH ($|E| = 18,703$) and the RH ($|E| = 19,613$), it represents a minimal fraction (approximately $1.16\%$) of the theoretical maximum distance for graphs of this order ($\sqrt{N(N-1)} = 18,731.5$). Further topological characterization shows that the SH effectively preserves the quality of RH, maintaining a density of $0.00011$ and an average degree of $2.00$ (compared to $0.00011$ and $2.09$ in the RH, respectively).

\begin{figure}[h]
\vspace{3mm}
\begin{center}
\includegraphics[width=1\textwidth]{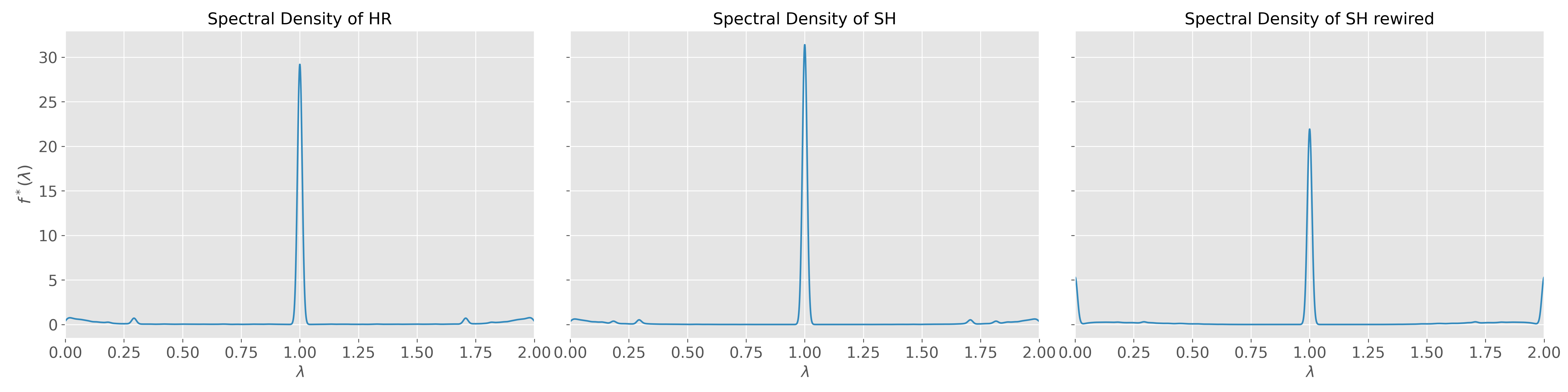}
\caption{\textbf{Laplacian spectral plots}.
Spectral density of the normalized graph Laplacian for three networks: (a) RH, (b) SH, (c) SH with all edges rewired. 
Densities are computed by convolving the eigenvalue spectrum with a Gaussian kernel ($\sigma$ = 0.01) and normalizing by the number of eigenvalues. All panels share the same y-axis to allow direct comparison. The Jensen–Shannon-based Laplacian spectral distance between RH and each variant is 0.085 (SH) and 0.224 (SH fully rewired).}
\end{center}
\vspace{-8mm}
\label{fig:spectrum}
\end{figure}

More importantly, the Laplacian Spectral Distance (LSD) calculated is $0.085$, indicating an high degree of global topological conservation against the normalized upper bound of $\ln(2) \approx 0.693$. This topological alignment is visually confirmed in Figure \ref{fig:spectrum}, where the spectral density of SH nearly overlaps with RH, particularly preserving the dominant peak at $\lambda = 1$. As noted by Banerjee and Jost~\cite{Banerjee_2007}, biological networks, such as metabolic and transcription networks, characteristically exhibit this distinctive spectral shape with a pronounced peak at $\lambda = 1$. 
Unlike generic random or scale-free models, this specific spectral profile indicates underlying vertex or motif duplications that fundamentally drive the evolution of these biological structures.

As illustrated in the spectral profiles in Figure~\ref{fig:spectrum}, the global rewiring results in a significant distortion of the density distribution and a suppression of the primary peak, characteristic of tree-like structure. The bootstrapping test yielded statistically significant $p$-value ($p < 0.005$), demonstrating that the observed similarity between the semantic reconstruction and the biological reference is not the  result of random chance but reflects a successful reconstruction of the underlying organizational logic of the Reactome database.

\section{Conclusion}
\label{sec:CONCLUSIONS}
In this work, we first obtained the manually curated biological hierarchy of Pathways and Reactions from Reactome related to \textit{Homo sapiens} by querying the Reactome Graph Database, which we named Reactome Hierarchy. Then, we extracted and concatenated the names and the description of the retrieved elements, and created the corresponding vector representations using SPECTER2. These embeddings served as the foundation to create a Semantic Hierarchy leveraging a combination of clustering and graph reconstruction algorithms that we designed.  
The two hierarchies were finally compared, and results indicate that the reconstructed Semantic Hierarchy can reasonably reproduce the reference Reactome structure, making a step towards bridging the gap between unstructured textual descriptions and structured biological information. 

This work represents the first attempt to check the feasibility and the accuracy of this workflow rather than an exhaustive study. The analysis was performed on a single species and relied on a limited set of comparison metrics, so the present findings should be read as a preliminary contribution within a broader research agenda. The current evaluation is largely computational and topological, while in future developments we plan to evaluate the results also from a biological perspective from a domain expert.

Future work include validating these findings with additional metrics that capture different aspects of graph similarity. Furthermore, the results are obtained exclusively from Reactome and the Homo sapiens hierarchy, and their generalisability to other species and biological databases remains to be assessed. Additionally, incorporating information like node-to-node similarity and graph alignment techniques could help better understand where the two hierarchies diverge. This type of information can be relevant in the identification of understudied/less curated areas, with a direct feedback to the curators. Taken together, these directions position the current results as a concrete step within a larger effort to bridge unstructured textual descriptions and structured biological knowledge, where the SH can be used to test and refine automatic curation tools which could greatly reduce the burden of manual curation, while preserving quality.

\section*{Conflict of interests}
\label{sec:CONFLICT-OF-INTERESTS}
The authors declare no conflict of interest.
  
\section*{Acknowledgments}
\label{sec:ACKNOWLEDGMENTS}
R.D.L. and S.B. are Ph.D. students enrolled in the National Ph.D. in AI, course on Health and Life Sciences, Università Campus Bio-Medico di Roma in collaboration with IAC-CNR. 


\section*{Availability of data and software code}
\label{sec:AVAILABILITY}
Our software code is available at the following URL: \href{https://baltig.cnr.it/sapiens/semhier}{https://baltig.cnr.it/sapiens/semhier}

\footnotesize
\bibliographystyle{unsrt}
\bibliography{bibliography_CIBB_file} 

\begin{thebibliography}{10}

\bibitem{ragueneau2026reactome}
Eliot Ragueneau, Chuqiao Gong, Pierre Sinquin, Cristoffer Sevilla, Deidre
  Beavers, Alexander Grentner, Johannes Griss, Gregory~FJ Hogue, Nancy~T Li,
  Lisa Matthews, et~al.
\newblock The reactome knowledgebase 2026.
\newblock {\em Nucleic Acids Research}, 54(D1):D673--D681, 2026.

\bibitem{fabregat2018reactomegraph}
Antonio Fabregat, Florian Korninger, Guilherme Viteri, Konstantinos
  Sidiropoulos, Pablo Marin-Garcia, Peipei Ping, Guanming Wu, Lincoln Stein,
  Peter D'Eustachio, and Henning Hermjakob.
\newblock Reactome graph database: Efficient access to complex pathway data.
\newblock {\em PLOS Computational Biology}, 14(1):e1005968, 2018.
\newblock Reactome Graph Database, release V95 (December 2025).

\bibitem{maeda2023automatic}
Kazuhiro Maeda and Hiroyuki Kurata.
\newblock Automatic generation of sbml kinetic models from natural language
  texts using gpt.
\newblock {\em International Journal of Molecular Sciences}, 24(8):7296, 2023.

\bibitem{wu2025application}
Guanming Wu, Lisa Matthews, Nathan Boyer, Marija Milacic, Deidre Beavers,
  Nancy~T Li, Bruce May, Karen Rothfels, Veronica Shamovsky, Ralf Stephan,
  et~al.
\newblock Application of large language models for annotating genes into
  reactome pathways.
\newblock {\em bioRxiv}, pages 2025--12, 2025.

\bibitem{mohammadi2025react}
Helia Mohammadi, Fatemeh Almodaresi, Gregory~FJ Hogue, Adam Wright, Marija
  Orlic-Milacic, Nancy~T Li, Amin Mawani, and Lincoln Stein.
\newblock React-to-me: A conversational interface for interactive exploration
  of the reactome pathway knowledgebase.
\newblock {\em bioRxiv}, pages 2025--12, 2025.

\bibitem{singh2023scirepevalmultiformatbenchmarkscientific}
Amanpreet Singh, Mike D'Arcy, Arman Cohan, Doug Downey, and Sergey Feldman.
\newblock Scirepeval: A multi-format benchmark for scientific document
  representations, 2023.

\bibitem{frohmann2024segmenttextuniversalapproach}
Markus Frohmann, Igor Sterner, Ivan Vulić, Benjamin Minixhofer, and Markus
  Schedl.
\newblock Segment any text: A universal approach for robust, efficient and
  adaptable sentence segmentation, 2024.

\bibitem{Wiley}
Leonard Kaufman and Peter~J. Rousseeuw.
\newblock {\em Agglomerative Nesting (Program AGNES)}, chapter~5, pages
  199--252.
\newblock John Wiley \& Sons, Ltd, 1990.

\bibitem{banerjee2009structuraldistanceevolutionaryrelationship}
Anirban Banerjee.
\newblock Structural distance and evolutionary relationship of networks, 2009.

\bibitem{Banerjee_2007}
Anirban Banerjee and Jürgen Jost.
\newblock Spectral plots and the representation and interpretation of
  biological data.
\newblock {\em Theory in Biosciences}, 126(1):15–21, March 2007.

\end{thebibliography}
\normalsize

\end{document}